\documentclass{article}

\usepackage[main, final]{neurips_2026}

\usepackage[utf8]{inputenc} 
\usepackage[T1]{fontenc}    
\usepackage{hyperref}       
\usepackage{url}            
\usepackage{booktabs}       
\usepackage{amsfonts}       
\usepackage{nicefrac}       
\usepackage{microtype}      
\usepackage{booktabs}       
\usepackage[table]{xcolor}  
\usepackage{array}          
\usepackage{tabularx}       
\usepackage{makecell}       
\usepackage{multirow}       
\usepackage{graphicx}
\usepackage{graphicx} 
\usepackage{algorithm}
\usepackage{algorithmic}
\usepackage{amsmath} 
\usepackage{enumitem}
\usepackage{tikz}
\definecolor{tablegray}{gray}{0.94}
\definecolor{tablegraylight}{gray}{0.97}

\newcolumntype{L}[1]{>{\raggedright\arraybackslash}p{#1}}
\newcolumntype{C}[1]{>{\centering\arraybackslash}p{#1}}
\newcolumntype{R}[1]{>{\raggedleft\arraybackslash}p{#1}}
\newcolumntype{Y}{>{\centering\arraybackslash}X}

\title{ValuSpec: Plug-and-Play Candidate Valuation before Target Verification for Tree-Based Speculative Decoding}

\author{%
Liang He\thanks{Email: \texttt{hel9919@163.com}} \\
Tongji University
\And
SiYuan Ma \\
Nanyang Technological University
\And
Qishi Zhan \\
Marquette University
\And
Yongqi Fan \\
Tencent
\AND
Mingyu Cao \\
University of Surrey
\And
Qizhen Lan \\
University of Texas Health Science Center at Houston
\And
Zhaolu Kang \\
Peking University
\And
Xilu Wang\thanks{Corresponding author: \texttt{wangxilu@surrey.ac.uk}} \\
University of Surrey
}

\begin{document}

\maketitle
\begin{abstract}
Tree-based speculative decoding accelerates large language model inference by verifying multiple candidate branches in parallel.   However, in practical serving scenarios, similar requests repeatedly induce overlapping candidate branches. 
While retrieval-based methods mitigate this by reusing historical drafts, the resulting candidate trees often contain many low-quality branches as the candidate tree grows, which reduces verification efficiency and wastes target-side computation. 
To address this, we propose ValuSpec, a plug-and-play candidate valuation mechanism that filters low-quality branches from a supplied candidate tree before expensive target-side verification. 
ValuSpec introduces transfer-based supervision to align filtering with the target model's behavior. 
A filtering threshold is estimated from the target model's greedy trajectories on a calibration split, without any parameter training. 
The target model then verifies the filtered tree under the output-identical greedy rule, so accepted tokens are guaranteed to match standard autoregressive decoding exactly.  
Experimental results on HumanEval, GSM8K, Dolly, and SpecBench show that, in our main  evaluation, ValuSpec achieves up to 2.38$\times$ end-to-end speedup in retrieval-based tree settings.
Furthermore, integrating ValuSpec into the EAGLE-3 dynamic tree framework yields up to 2.78$\times$ end-to-end speedup without modifying its tree construction strategy, demonstrating that ValuSpec serves as a generally applicable and plug-and-play candidate valuation module.
\end{abstract}

\section{Introduction}
\label{sec:intro}

Large language models (LLMs) are increasingly deployed in real-world serving systems, where inference efficiency directly impacts latency and system scalability~\citep{Zhou2024EfficientLLM,Park2025InferenceSurvey}. In such environments, requests are not independent at the level of local branches. Different users often submit similar programming queries, reasoning tasks, or instruction-following prompts, which induce overlapping phrases, code fragments, or reasoning steps during decoding~\citep{Potamitis2025CacheSaver,Zhang2025RelServe}.
Despite this structure, modern LLM serving still relies on autoregressive (AR) decoding, where tokens are generated sequentially and each step requires a full forward pass through the model~\citep{brown2020gpt3,touvron2023llama}. This sequential dependency limits parallelism and leads to high latency, especially for long outputs~\citep{khoshnoodi2024survey,miao2025efficientserving}.

\begin{figure*}[h]
  \centering
  \includegraphics[width=1\textwidth]{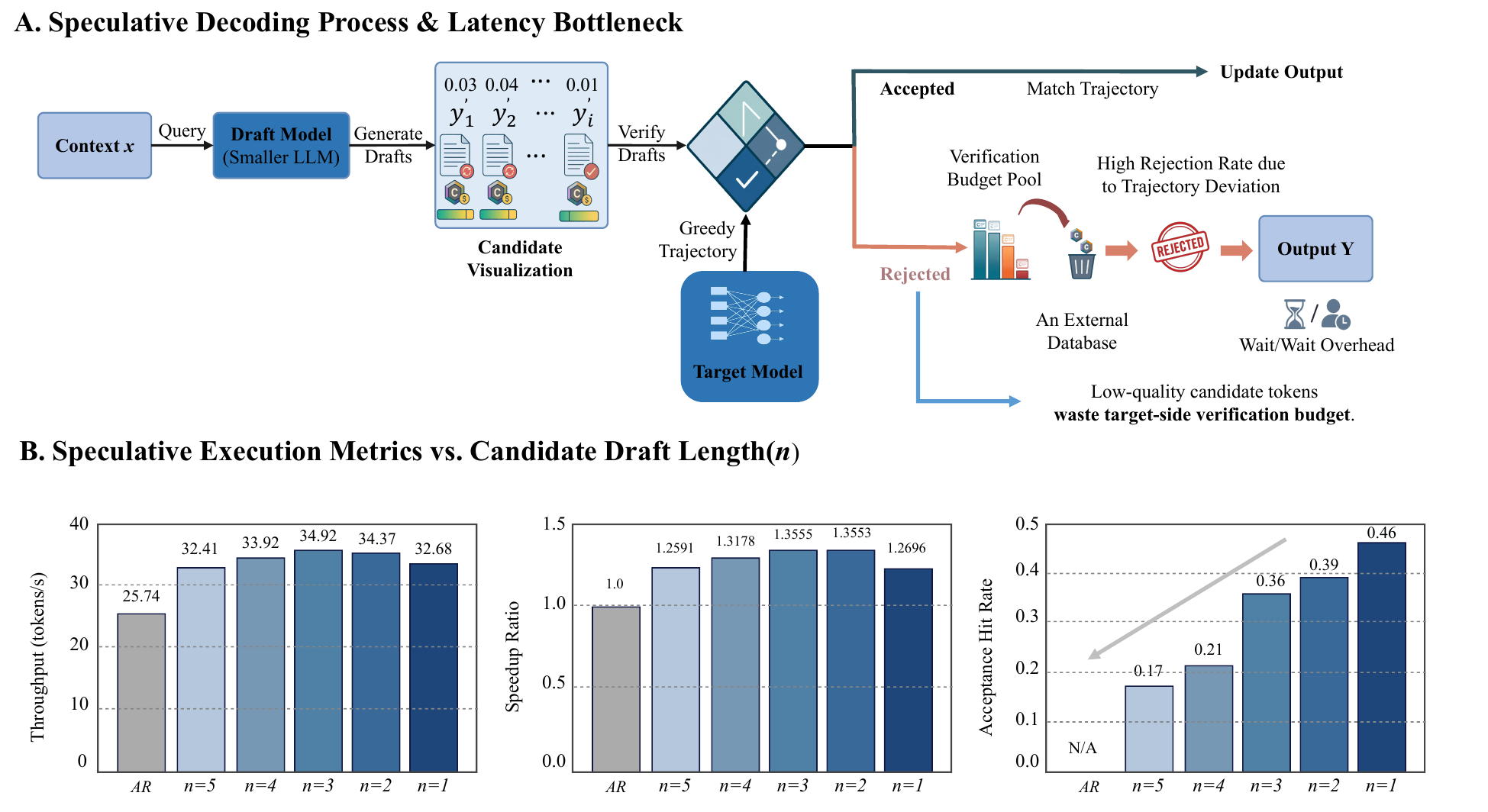}
  \caption{
  \textbf{Candidate quality and verification efficiency.}
  (A) Low-quality candidate tokens waste target-side verification budget in SD. Candidate tokens that deviate from the target model's greedy trajectory are rejected, yet still consume verification computation without contributing to decoding progress.
  (B) Acceptance rate versus candidate draft length on Dolly (A100) using Llama-2-7b-chat-hf. Acceptance decreases monotonically as the candidate draft length grows, indicating increasing redundancy and a higher proportion of low-quality candidates.
  }
  \label{fig:candidate_quality}
  \vspace{-0.2cm}
\end{figure*}

Speculative decoding (SD) has been proposed to accelerate AR decoding while preserving output quality~\citep{leviathan2023fast,chen2023acceleratinglargelanguagemodel,xia2024unlocking}. It follows a draft--verify paradigm, where a lightweight draft model  proposes candidate tokens and a larger target model verifies them in parallel~\citep{leviathan2023fast,chen2023acceleratinglargelanguagemodel,liu2023online}. When multiple tokens are accepted in a single iteration, the target model can advance by several steps at once, improving throughput~\citep{leviathan2023fast,chen2023acceleratinglargelanguagemodel,yan2025decoding}. However, the efficiency of SD fundamentally relies on the quality of candidate tokens~\citep{liu2023online,huang2024specdec}. As illustrated in Figure~\ref{fig:candidate_quality} A, many candidate tokens forwarded to the target model deviate from its greedy trajectory and are rejected early~\citep{liu2023online}. These low-quality candidates still incur target-side verification cost, but do not contribute to accepted decoding progress~\citep{huang2024specdec}. As a result, substantial target-side computation is wasted on candidates that will be rejected~\citep{zhou2026calibrated,du2026reinforcement}.

In practical LLM serving, this inefficiency is further amplified by repeated computation at the draft stage~\citep{jin2024batchllm,gao2025context,kim2025rethinking}. Since many requests share similar local continuation patterns, the draft model repeatedly regenerates highly overlapping candidate fragments across requests~\citep{pan2025marconi,hu2025deepsserve}. This redundancy occurs even before verification begins.
Retrieval-based tree speculative decoding addresses this issue by reusing historical drafts instead of regenerating them online~\citep{he2024rest}. Candidate branches are stored in an external database and retrieved to construct a candidate tree during inference. This approach is particularly effective in recurring workloads, where future requests often resemble past queries.
However, retrieval introduces a new challenge. As the database grows, the retrieved candidate tree becomes increasingly redundant~\citep{he2024rest}. Many candidate sequences may match the current context superficially but diverge early from the target model's behavior~\citep{leviathan2023fast,liu2023online}. These low-quality candidates are still forwarded to the target model and rejected during verification, consuming computation without contributing useful decoding progress~\citep{huang2024specdec}.

This redundancy is reflected in the acceptance dynamics. As shown in Figure~\ref{fig:candidate_quality} B (A detailed sensitivity analysis can be found in Appendix \ref{sec:sensitivity_draft}), increasing the size $n$ of the candidate draft length leads to a monotonic decrease in acceptance rate~\citep{xia2024unlocking,yan2025decoding}. Larger candidate sets are more likely to include candidates that deviate from the target model's trajectory, resulting in more rejected tokens and less effective progress per iteration~\citep{liu2023online}. These observations reveal a shift in the system bottleneck. Once candidate reuse is introduced, the key challenge is no longer how to generate candidate tokens, but how to determine which candidates are worth verifying~\citep{zhou2026calibrated}.

To address this problem, we propose \textit{ValuSpec}, a plug-and-play candidate valuation mechanism for tree-based speculative decoding. Rather than modifying candidate construction, ValuSpec operates after a candidate tree has already been constructed and before target-side verification, filtering low-value branches from the supplied tree. This positioning distinguishes ValuSpec from generation-side tree optimization or pruning methods, which modify how candidate trees are constructed, expanded, or reduced during candidate supply. The two therefore operate at different stages of the speculative-decoding pipeline and can be applied sequentially.

To align candidate valuation with the target model, ValuSpec introduces a transfer-based supervision strategy. Specifically, deterministic autoregressive trajectories from the target model are used to calibrate a filtering threshold, without requiring additional parameter training. The filtered candidate tree is then verified under the standard output-identical greedy rule, ensuring that the final output remains the same as deterministic autoregressive decoding. Although ValuSpec is primarily motivated by retrieval-based tree settings, it is not tied to a specific candidate supplier: once a candidate tree is materialized, the same verifier-side valuation module can be applied before target verification. We therefore focus on retrieval-based tree speculative decoding as the primary setting and additionally evaluate ValuSpec on top of dynamically constructed and generation-side optimized candidate trees.

Our contributions are summarized as follows:
\begin{itemize}[leftmargin=1.5em, itemsep=0.4em]
\item We identify candidate valuation before target verification as a distinct bottleneck in tree-based speculative decoding for practical LLM serving. In particular, while retrieval-based methods alleviate repeated online draft generation by reusing historical drafts, the resulting candidate trees often contain many low-quality branches that waste target-side verification budget.

\item We propose ValuSpec, a plug-and-play verifier-side candidate valuation mechanism for tree-based speculative decoding. Unlike generation-side methods that optimize tree construction or pruning during candidate supply, ValuSpec operates after the candidate tree has been materialized and before target verification, allowing it to be composed with different upstream candidate suppliers without modifying their generation procedures.

\item We introduce transfer-based supervision, where deterministic autoregressive trajectories from the target model are used to guide the lightweight plug-and-play candidate valuation model  for branch pruning while preserving exact output equivalence.

\item Experimental results show that ValuSpec improves acceptance-related efficiency, reduces verified tree size, and increases end-to-end throughput across different candidate suppliers. In addition to retrieval-based trees, ValuSpec provides substantial additional speedup when applied on top of generation-side optimized methods, supporting the complementarity between candidate supply optimization and verifier-side candidate valuation.
\end{itemize}

\section{Related Work}
\label{sec:related_work}
\textbf{Speculative Decoding and Candidate Supply.}
Speculative decoding accelerates autoregressive inference by allowing a lightweight draft model  to generate candidate tokens that are then verified by a larger target model~\citep{leviathan2023fast,chen2023acceleratinglargelanguagemodel,xia2024unlocking}. 
A major line of work focuses on improving candidate supply, i.e., producing candidate tokens or trees that better match the target model~\citep{yan2025decoding,huang2024specdec}. This includes tree construction methods that shape candidate trees during generation, as well as drafter-side training approaches such as BudgetDraft~\citep{budgetdraft}, which exposes the drafter to multiple sparse KV views to recover acceptance rates under memory constraints.
Retrieval-based methods, such as REST~\citep{he2024rest}, reuse historical branches from an external database to construct candidate trees, reducing repeated online generation under recurring workloads. While effective for candidate reuse, these approaches often produce large candidate trees with substantial redundancy, leaving verifier-side inefficiency largely unaddressed~\citep{zhou2026calibrated}.
ValuSpec complements this line of work by focusing on candidate valuation rather than candidate supply. Instead of constructing better trees, it filters low-quality candidates from a supplied tree before target-side verification, and can be applied on top of retrieval-based suppliers.

\textbf{Dynamic Tree Construction and Generation-Side Pruning.}
Another line of work improves speculative decoding by optimizing candidate-tree construction, expansion, or pruning on the generation side~\citep{xia2024unlocking,yan2025decoding}. Methods such as EAGLE-3~\citep{li2025eagle3scalinginferenceacceleration} dynamically construct candidate trees to better align candidate supply with the target model~\citep{wang2026talon}. Other generation-side pruning methods similarly reduce low-value candidates while constructing or expanding the candidate tree.
ValuSpec operates at a different stage. It takes an already materialized candidate tree as input and performs candidate valuation immediately before target verification. Thus, generation-side pruning determines which candidates are supplied, whereas ValuSpec determines which of the supplied candidates are worth sending to the target verifier. The two mechanisms are therefore complementary rather than mutually exclusive. Our experiments confirm that ValuSpec provides additional acceleration when applied on top of generation-side optimized candidate suppliers.

\textbf{Verifier-Side Candidate Valuation.}
Compared with candidate supply, verifier-side candidate valuation has received less attention. Both retrieval-based and dynamic-tree methods can still pass low-quality candidates to the target model, leading to wasted verification~\citep{liu2023online,huang2024specdec}.
ValuSpec explicitly targets this problem by introducing a plug-and-play valuation mechanism that filters candidates before verification using transfer-based supervision, reducing verifier-side redundancy while preserving exact output equivalence.
\begin{figure*}[t]
  \centering
  \includegraphics[width=1\textwidth]{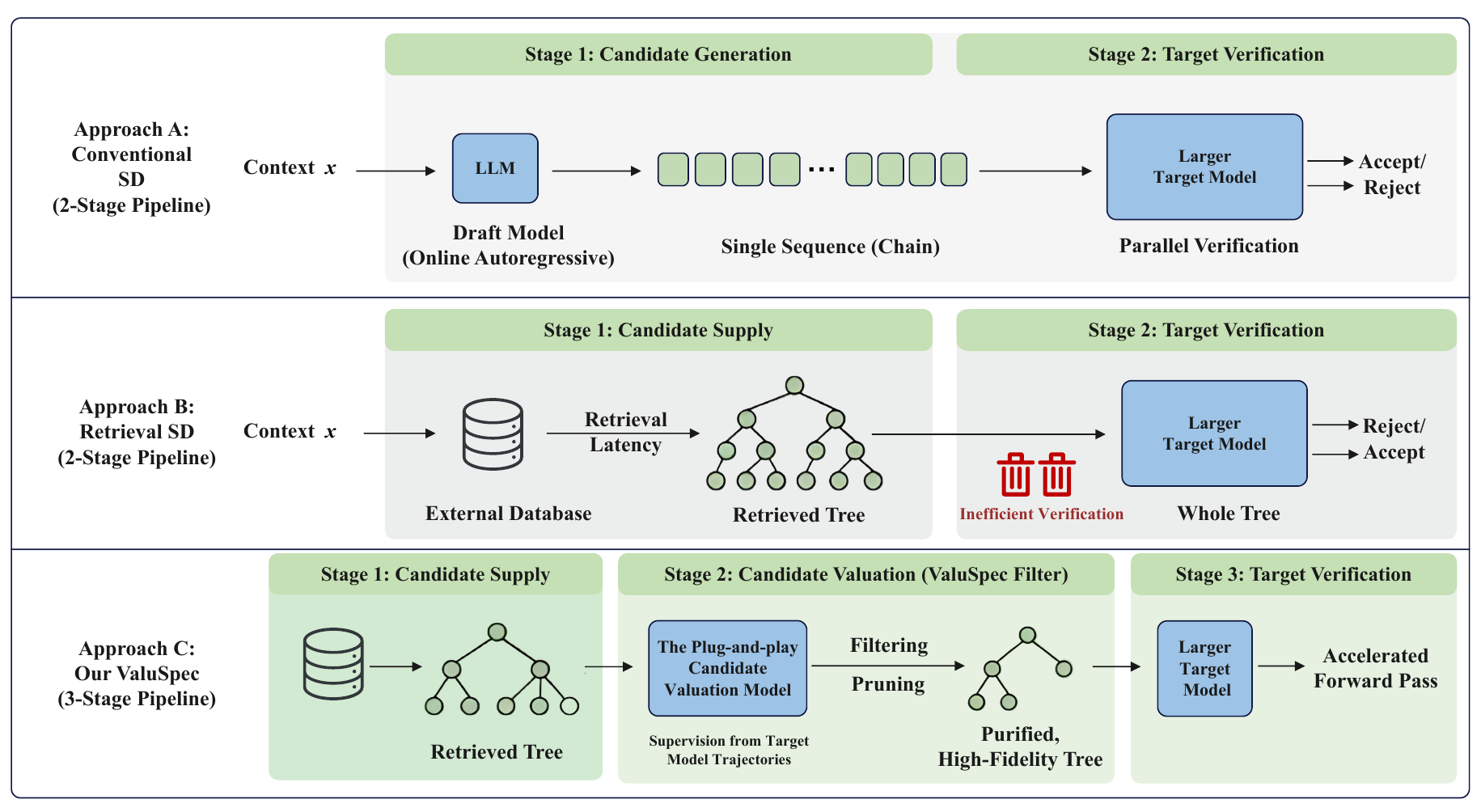}
  \caption{
  \textbf{Comparison of speculative decoding paradigms.} 
  Conventional speculative decoding generates candidates online for each request, leading to repeated computation under similar inputs. 
  Retrieval-based speculative decoding reuses historical candidates but introduces large candidate trees with many low-quality candidates that waste target-side verification. 
  ValuSpec is a plug-and-play candidate valuation mechanism to filter low-quality candidates from the supplied tree before target-side verification, reducing verification redundancy without modifying candidate construction.
  }
  \label{fig:pipeline_compare}
  \vspace{-0.2cm}
\end{figure*}

\section{ValuSpec: Plug-and-Play Candidate Valuation before Target Verification}
\label{sec:ValuSpec}

\textbf{Problem formulation.}
Let $\mathcal{V}$ be the vocabulary. Given a context $x=\left(x^1, \ldots, x^t\right) \in \mathcal{V}^{\otimes t}$, the target language model $p_{\theta}$ predicts the probability of the next word $p_{\theta}(\cdot\mid x)$, from which we sample the next token $x^{t+1}$. We focus on deterministic greedy decoding, where the target autoregressive decoder selects:
\begin{equation}
y \;=\; \arg\max_{y'\in\mathcal{V}} p_{\theta}(y'\mid x).
\label{eq:greedy_accept}
\end{equation}
At position $i$ under prefix $(x,y_{<i})$, the corresponding greedy decision is:
\begin{equation}
y_i^*(x) \;=\; \arg\max_{y\in\mathcal{V}} p_{\theta}\!\left(y \mid x,y_{<i}\right).
\label{eq:greedy_step_i}
\end{equation}
Starting from the context $x$, greedy autoregressive decoding repeatedly applies Eq.~\eqref{eq:greedy_step_i} to generate the reference output sequence $Y^{*}(x)$. Our goal is to reduce end-to-end decoding time while preserving output identity. Formally, for every context $x$, ValuSpec guarantees to return the same output sequence as greedy autoregressive decoding under $p_{\theta}$, i.e., $Y(x)=Y^{*}(x)$, while improving runtime efficiency through candidate valuation before target-side verification. Therefore, any token accepted by ValuSpec must satisfy Eq.~\eqref{eq:greedy_step_i} under its corresponding prefix context. A comparison between the two representative SD paradigms and our ValuSpec is presented in Figure~\ref{fig:pipeline_compare}.

\begin{figure*}[h]
  \centering
  \includegraphics[width=0.98\textwidth]{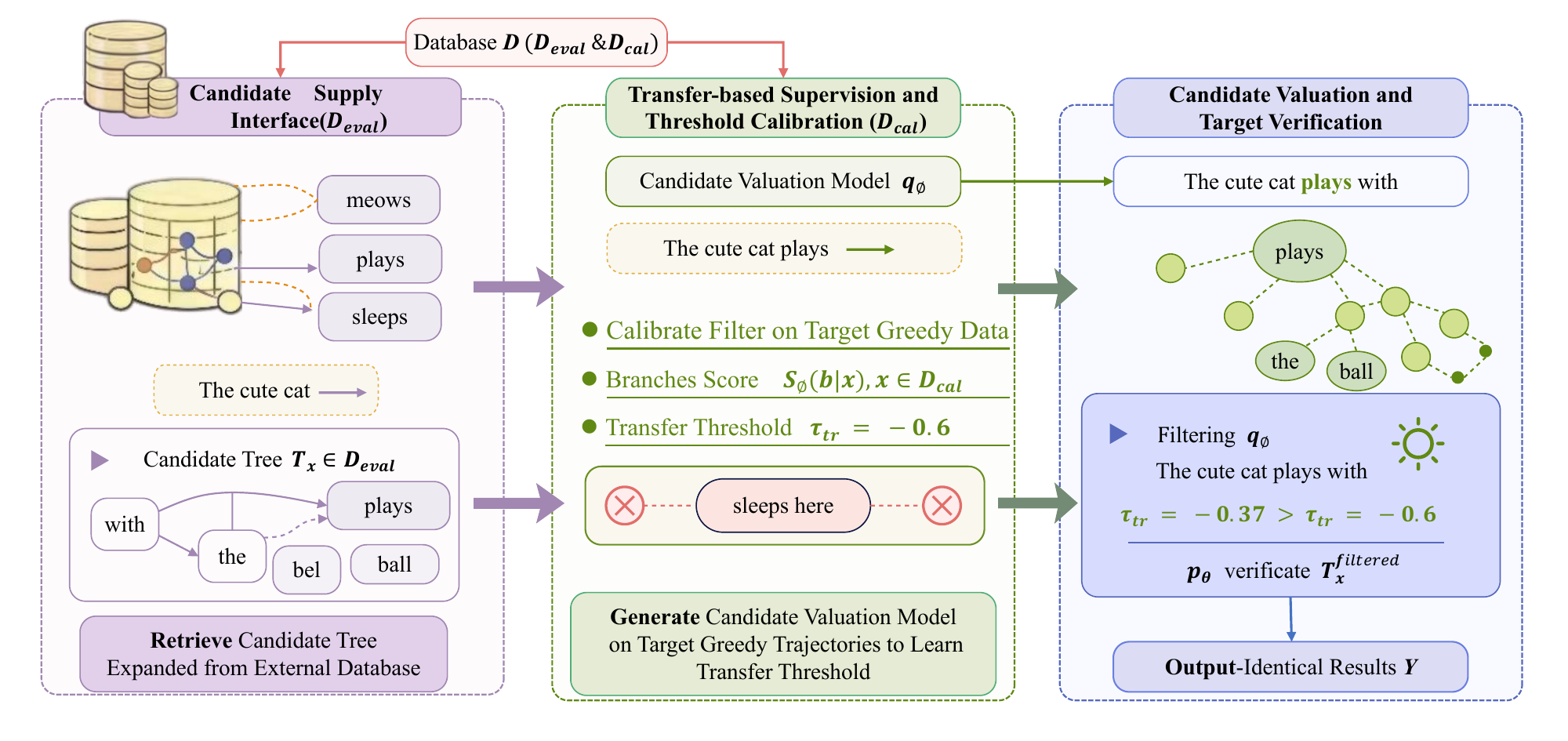}
  \caption{\textbf{Overview of ValuSpec.} ValuSpec decouples candidate supply from candidate valuation using two disjoint splits of the same database \(\mathcal{D}\). A candidate tree \(T_x\) is supplied from \(\mathcal{D}_{\mathrm{eval}}\), while a lightweight draft model  \(q_\phi\) is calibrated on target greedy trajectories from \(\mathcal{D}_{\mathrm{cal}}\) to filter low-quality branches before target-side verification. The target model \(p_\theta\) then verifies the filtered subtree \(T_x^{\mathrm{filtered}}\) with tree masking/attention and accepts the longest greedy-consistent prefix, producing the final output sequence \(Y\) under output-identical decoding.}
  \label{fig:Figure3}
\end{figure*}

\textbf{Overview.}
To reduce wasted target-side verification, ValuSpec decouples candidate supply from candidate valuation, as illustrated in Figure~\ref{fig:Figure3}. Given a candidate tree $T_x$ for context $x$, ValuSpec introduces a plug-and-play candidate valuation model $q_{\phi}$ to score each branch $b \in T_x$ and filter low-quality branches before expensive target-side verification. We denote by $s_{\phi}(b\mid x)$ the score assigned by the candidate valuation model $q_{\phi}$ to a candidate branch $b$ under context $x$. 
ValuSpec consists of three stages.
First, a candidate tree $T_x$ is supplied by an upstream tree supplier. In the primary setting studied in this paper, $T_x$ is instantiated by a retrieval-based tree supplier built from an external database; in the dynamic-tree setting, $T_x$ is supplied by EAGLE-3. Second, ValuSpec calibrates a lightweight draft model  $q_{\phi}$ using deterministic target greedy trajectories and derives a transfer threshold $\tau_{\mathrm{tr}}$. Third, at inference time the calibrated candidate valuation model is used to remove low-quality branches, yielding a compact subtree $T_x^{\mathrm{filtered}}$. Finally, the target model performs tree-MCSD verification on $T_x^{\mathrm{filtered}}$ and accepts the longest prefix consistent with greedy decoding, advancing multiple tokens with a single forward pass. Throughout, we enforce output-identical decoding under deterministic autoregression, so ValuSpec’s final output exactly matches standard target-model decoding. See the pseudocode of ValuSpec in Appendix~\ref{app:full_algorithm}.

\textbf{Candidate supply interface.}
Since ValuSpec operates as a plug-and-play module, it takes as input a candidate tree $T_x$ that has already been materialized before target-side verification, regardless of how it was constructed. This design makes ValuSpec compatible with different upstream candidate suppliers.
In the primary retrieval-based setting, $T_x$ is instantiated from an external database or prefix trie under a fixed token budget. In the dynamic-tree setting, $T_x$ is supplied by a dynamic tree construction method such as EAGLE-3. Once the candidate tree is available, ValuSpec applies the same filtering mechanism before target-side verification.

\textbf{Transfer-based supervision and threshold calibration.}
A key challenge in candidate valuation is determining which branches are likely to be accepted by the target model. Heuristic confidence scores such as entropy or top-1 probability for pruning are not directly tied to the target model's behavior and therefore provide unreliable filtering signals. Instead, ValuSpec derives a filtering threshold directly from target-generated greedy trajectories on a calibration split $\mathcal{D}_{\mathrm{cal}}$, which represent the paths the target model would naturally follow. 

To achieve this, on a calibration split $\mathcal{D}_{\mathrm{cal}}$, we run the target model with greedy decoding to obtain trajectories $b^{*}(x) = (y^{*}_1, \ldots, y^{*}_{\ell})$ for each context $x$, where $y^{*}_i = \arg\max_{y} p_{\theta}(y \mid x, y^{*}_{<i})$. The candidate valuation model $q_{\phi}$ is used as a fixed scoring model without any parameter updates. 
Given a candidate branch $b=(y_1,\ldots,y_{\ell})$ under context $x$, we define a sequence-level score by the mean log-probability under the lightweight plug-and-play candidate valuation model:
\begin{equation}
s_{\phi}(b \mid x) = \frac{1}{\ell} \sum_{i=1}^{\ell} \log q_{\phi}(y_i \mid x, y_{<i}).
\label{Eq:Score}
\end{equation}
We compute $s_{\phi}(b^{*}(x)\mid x)$ for all $x \in \mathcal{D}_{\mathrm{cal}}$ and set the transfer threshold as a lower quantile of this score distribution:
\begin{equation}
\tau_{\mathrm{tr}} = \mathrm{Quantile}_{\delta}\!\left(\left\{\, s_{\phi}(b^{*}(x) \mid x) : x \in \mathcal{D}_{\mathrm{cal}} \,\right\}\right).
\end{equation}
The threshold $\tau_{\mathrm{tr}}$ is estimated from calibration data rather than manually tuned; the only hyperparameter is the quantile level $\delta$, which controls filtering conservativeness (A detailed analysis can be found in Appendix \ref{sec:sensitivity_quantile}). At inference time, ValuSpec uses $\tau_{\mathrm{tr}}$ to filter branches whose scores are lower than $\tau_{\mathrm{tr}}$, yielding a target-aligned branch filtering rule for online tree-based speculative decoding. Note that we use $\mathcal{D}_{\mathrm{cal}}$ only to (i) generate target greedy trajectories and (ii) estimate the transfer threshold $\tau_{\mathrm{tr}}$. All throughput and speedup results are measured on a separate evaluation split $\mathcal{D}_{\mathrm{eval}} \subset \mathcal{D}$ with $\mathcal{D}_{\mathrm{cal}} \cap \mathcal{D}_{\mathrm{eval}} = \emptyset$.

\textbf{Candidate valuation and target verification.}
At inference time, given a candidate tree $T_x$ supplied before target-side verification, we compute $s_{\phi}(b \mid x)$ via Eq.~\eqref{Eq:Score} for each branch $b \in T_x$ and retain only branches that meet the transfer threshold $\tau_{\mathrm{tr}}$:
\begin{equation}
T_x^{\mathrm{filtered}} \;=\; \{\, b \in T_x \;:\; s_{\phi}(b \mid x) \ge \tau_{\mathrm{tr}} \,\}.
\label{eq:prune_rule}
\end{equation}
This filtering step concentrates target-side computation on branches in $T_x^{\mathrm{filtered}}$ that are more consistent with the target model’s greedy behavior, reducing wasted verification. If $T_x^{\mathrm{filtered}} = \emptyset$, we fall back to one-step greedy autoregressive decoding with the target model.
The filtered subtree $T_x^{\mathrm{filtered}}$ is then verified in a single target forward pass using tree attention or masking, yielding $p_{\theta}(\cdot\mid x,y_{<i})$ for all nodes simultaneously. The output-identical greedy rule in Eq.~\eqref{eq:greedy_step_i} is applied along each branch and accepts the longest valid prefix $\Delta$. If $\Delta \neq [\,]$, we append accepted tokens to both the output and the context, i.e., $Y \leftarrow Y \oplus \Delta$ and $x \leftarrow x \oplus \Delta$. If no token is accepted ($\Delta=[\,]$), we fall back to one-step greedy autoregressive decoding with the target model and set $\hat y$ by Eq.~\eqref{eq:greedy_accept}. We then update $Y \leftarrow Y \oplus \hat y$ and $x \leftarrow x \oplus \hat y$, and proceed to the next iteration until EOS, returning the final output $Y$.

\section{Experimental Setup}
\label{sec:exp_setup}

\textbf{Benchmarks.}
We evaluate ValuSpec on three representative generation benchmarks: HumanEval for code generation (164 Python problems)~\citep{Chen2021EvaluatingLL}, GSM8K for arithmetic reasoning~\citep{cobbe2021training}, and Dolly for instruction following\footnote{\url{https://huggingface.co/datasets/databricks/databricks-dolly-15k}}. These workloads cover diverse decoding patterns, including structured code generation, reasoning-heavy tasks, and open-ended instruction responses. 
For the EAGLE-3 dynamic-tree integration experiment, we additionally add SpecBench\footnote{\url{https://github.com/hemingkx/Spec-Bench}} to evaluate candidates on a multi-task SD benchmark.
All experiments use greedy decoding (temperature $T=0$) to ensure output-identical comparisons across methods. Unless otherwise stated, the main experiments use batch size 1; we additionally evaluate batch sizes 4 and 8 to assess ValuSpec under batched inference.

\textbf{Recurring-workload protocol.}
To simulate realistic serving conditions, we adopt a non-overlapping split of each dataset into a calibration set $\mathcal{D}_{\mathrm{cal}}$ (70\%) and an evaluation set $\mathcal{D}_{\mathrm{eval}}$ (30\%), with $\mathcal{D}_{\mathrm{cal}} \cap \mathcal{D}_{\mathrm{eval}} = \emptyset$. 
The calibration set $\mathcal{D}_{\mathrm{cal}}$ is used exclusively to compute target greedy trajectories and estimate the transfer threshold $\tau_{\mathrm{tr}}$. The evaluation set $\mathcal{D}_{\mathrm{eval}}$ is used for inference and throughput measurement. Within $\mathcal{D}_{\mathrm{eval}}$, we further ensure that candidate-tree construction and evaluation contexts are drawn from disjoint subsets, so that no evaluation query is used to build its own candidate tree.
This protocol simulates a recurring-workload setting in which historical requests (used for calibration) and future requests (used for evaluation) share similar local continuation patterns, allowing threshold reuse without online adaptation.

\textbf{Cross-dataset transfer.}
To evaluate robustness under workload shift, we additionally perform cross-dataset experiments. In this setting, the threshold $\tau_{\mathrm{tr}}$ is calibrated on one dataset and directly applied to another dataset without re-calibration. This tests whether the learned filtering rule generalizes beyond the calibration workload.

\textbf{Models.}
We use Llama-2-7b-chat-hf and Llama-2-13b-chat-hf as target models ($p_{\theta}$), and JackFram/LLaMA-68M\footnote{\url{https://huggingface.co/JackFram/llama-68m}} as the lightweight plug-and-play candidate valuation model ($q_{\phi}$). The valuation model shares the same tokenizer and architecture family as the target models, enabling efficient sequence scoring without vocabulary mismatch. All models are loaded in FP16 precision.
For dynamic-tree experiments, we use Meta-Llama/Llama-3.1-8B-Instruct~\citep{dubey2024llama3} as the target model in combination with EAGLE-3 for candidate tree construction.
For the additional Qwen3.5 experiments, we use Qwen3.5-9B as the target model and Qwen3.5-0.8B as the lightweight candidate valuation model; in the Standard SD setting, Qwen3.5-0.8B also serves as the draft model.

\textbf{Candidate tree construction.}
ValuSpec operates on candidate trees supplied before target-side verification and does not depend on a specific tree construction mechanism. In our experiments, we consider two representative candidate suppliers: Retrieval-based trees, where candidate trees are constructed from an external database under a fixed draft-token budget (64 tokens)~\citep{he2024rest}; Dynamically constructed trees, where candidate trees are generated online by EAGLE-3 with a larger effective budget (approximately 256 tokens)~\citep{li2025eagle3scalinginferenceacceleration}. This setup allows us to evaluate ValuSpec both in its primary retrieval-based setting and in a dynamic-tree setting.

\textbf{Compared methods.}
ValuSpec is compared with four baselines: AR (greedy autoregressive decoding), SD (output-identical SD), Retrieval SD (retrieval-based tree verification without pre-filtering, draft tokens $=64$)~\citep{he2024rest}, and EAGLE-3 (dynamic tree construction before target-side verification)~\citep{li2025eagle3scalinginferenceacceleration}.
For EAGLE-3, we additionally evaluate the integration of ValuSpec as a plug-and-play filtering module applied after dynamic tree construction.
All methods enforce the same output-identical constraint, requiring that accepted tokens match the greedy decision of the target model.

\textbf{Metrics and implementation.}
We report end-to-end decoding throughput (tokens/s) and speedup over AR on the same hardware, plus acceptance rate and filtered tree size for verifier-side efficiency.
All main experiments are conducted on a single NVIDIA A100 GPU. To assess generalisability for real-world deployment, we also evaluate on an RTX 4090, with the corresponding results presented in Appendix~\ref{SM B: 4090}. We implement all methods using PyTorch and the HuggingFace Transformers library, and adopt standard tree-based parallel verification with masking/attention for candidate trees.
\section{Experimental Results}
\label{sec:exp_results}

In this section, we evaluate ValuSpec as a verifier-side candidate valuation module operating after candidate generation and immediately before target verification. Throughout all experiments, the upstream candidate-generation, retrieval, and tree-construction procedures remain unchanged; ValuSpec only determines which already generated branches are forwarded to the target verifier. We first evaluate retrieval-supplied trees under recurring workloads, then study cross-dataset threshold transfer and integration with dynamically constructed EAGLE-3 trees. Finally, we provide additional validation across candidate suppliers, model families, filtering rules, and batch sizes.

\subsection{Main Results: Recurring-Workload Evaluation}
\label{sec:recurring_eval}

We first evaluate ValuSpec on retrieval-supplied candidate trees under recurring workloads. In this setting, the retrieval mechanism first constructs the candidate tree, after which ValuSpec is applied only before target verification. For each dataset, 70\% of the data is used to calibrate the transfer threshold $\tau_{\mathrm{tr}}$, and the remaining 30\% is used for inference without further tuning.
\begin{figure*}[b]
    \centering
    \includegraphics[width=\linewidth, trim=1cm 1.15cm 0 0, clip]{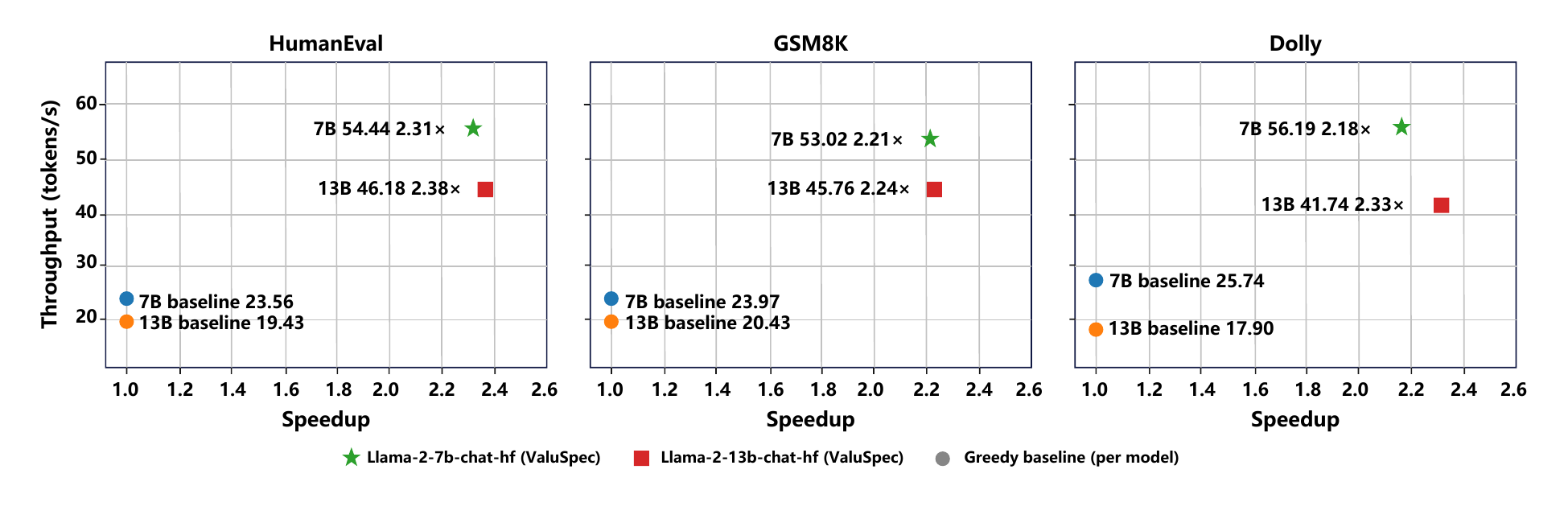}
    \caption{Main results of ValuSpec under output-identical decoding across three decoding loads. ValuSpec uses $n=4$, draft tokens $=64$, and $\delta=0.05$ for all tasks.}
    \label{fig:ValuSpec_main_results_scatter}
\end{figure*}
Table~\ref{tab:recurring_results} reports the main results. Across all three workloads and both target model sizes, ValuSpec consistently achieves the highest throughput while preserving exact output equivalence. On the 7B target, ValuSpec achieves 2.18$\times$--2.31$\times$ speedup over AR; on the 13B target, it achieves 2.24$\times$--2.38$\times$ speedup. Figure~\ref{fig:ValuSpec_main_results_scatter} shows the same consistent improvement across workloads.
Importantly, the Retrieval SD candidate supplier is kept unchanged: it first constructs the candidate tree, and ValuSpec subsequently filters low-value branches before target verification. The additional gain therefore comes from reducing residual verifier-side computation rather than modifying candidate generation or retrieval. This distinction suggests that candidate supply and candidate valuation constitute separate optimization stages.

\begin{table}[t]
\centering
\caption{\textbf{Recurring-workload evaluation.}
Throughput (tokens/s) under output-identical decoding on A100, with speedup over AR shown in parentheses.
SD uses $n=3$; Retrieval SD uses a REST-style tree with 64 draft tokens; ValuSpec uses $n=4$, 64 draft tokens, and $\delta=0.05$.
}
\label{tab:recurring_results}

\small
\setlength{\tabcolsep}{3.8pt}
\renewcommand{\arraystretch}{1.18}

\begin{tabular}{@{}lcccccccc@{}}
\toprule
& \multicolumn{4}{c}{\textbf{Llama-2-7B}}
& \multicolumn{4}{c}{\textbf{Llama-2-13B}} \\
\cmidrule(lr){2-5}
\cmidrule(lr){6-9}

\textbf{Task}
& \textbf{AR}
& \makecell{\textbf{SD}\\[-1pt]\textbf{($n=3$)}}
& \makecell{\textbf{Retrieval}\\[-1pt]\textbf{SD}}
& \cellcolor{tablegraylight}\textbf{ValuSpec}
& \textbf{AR}
& \makecell{\textbf{SD}\\[-1pt]\textbf{($n=3$)}}
& \makecell{\textbf{Retrieval}\\[-1pt]\textbf{SD}}
& \cellcolor{tablegraylight}\textbf{ValuSpec} \\
\midrule

HumanEval-x-python
& \makecell{23.56\\{\footnotesize (1.00$\times$)}}
& \makecell{29.20\\{\footnotesize (1.24$\times$)}}
& \makecell{40.64\\{\footnotesize (1.72$\times$)}}
& \cellcolor{tablegraylight}\makecell{
\textbf{54.44}\\
{\footnotesize \textbf{(2.31$\times$)}}
}
& \makecell{19.43\\{\footnotesize (1.00$\times$)}}
& \makecell{25.27\\{\footnotesize (1.30$\times$)}}
& \makecell{33.61\\{\footnotesize (1.73$\times$)}}
& \cellcolor{tablegraylight}\makecell{
\textbf{46.18}\\
{\footnotesize \textbf{(2.38$\times$)}}
}
\\

\addlinespace[2pt]

GSM8K-main-test
& \makecell{23.97\\{\footnotesize (1.00$\times$)}}
& \makecell{32.07\\{\footnotesize (1.34$\times$)}}
& \makecell{41.36\\{\footnotesize (1.73$\times$)}}
& \cellcolor{tablegraylight}\makecell{
\textbf{53.02}\\
{\footnotesize \textbf{(2.21$\times$)}}
}
& \makecell{20.43\\{\footnotesize (1.00$\times$)}}
& \makecell{28.32\\{\footnotesize (1.39$\times$)}}
& \makecell{36.99\\{\footnotesize (1.81$\times$)}}
& \cellcolor{tablegraylight}\makecell{
\textbf{45.76}\\
{\footnotesize \textbf{(2.24$\times$)}}
}
\\

\addlinespace[2pt]

Dolly
& \makecell{25.74\\{\footnotesize (1.00$\times$)}}
& \makecell{34.89\\{\footnotesize (1.36$\times$)}}
& \makecell{45.68\\{\footnotesize (1.77$\times$)}}
& \cellcolor{tablegraylight}\makecell{
\textbf{56.19}\\
{\footnotesize \textbf{(2.18$\times$)}}
}
& \makecell{17.90\\{\footnotesize (1.00$\times$)}}
& \makecell{26.99\\{\footnotesize (1.51$\times$)}}
& \makecell{34.73\\{\footnotesize (1.94$\times$)}}
& \cellcolor{tablegraylight}\makecell{
\textbf{41.74}\\
{\footnotesize \textbf{(2.33$\times$)}}
}
\\

\bottomrule
\end{tabular}
\end{table}

\subsection{Cross-Dataset Threshold Transfer}
\label{sec:cross_dataset_eval}

\begin{table}[t]
\centering
\caption{\textbf{Cross-dataset threshold transfer.}
The threshold is calibrated on the 70\% split of one dataset and transferred to another without re-calibration.
We report throughput (tokens/s), with speedup over AR shown in parentheses.
}
\label{tab:cross_dataset_results}

\small
\setlength{\tabcolsep}{4.2pt}
\renewcommand{\arraystretch}{1.18}

\begin{tabular}{@{}llcccc@{}}
\toprule
&
& \multicolumn{2}{c}{\textbf{Llama-2-7B}}
& \multicolumn{2}{c}{\textbf{Llama-2-13B}} \\
\cmidrule(lr){3-4}
\cmidrule(lr){5-6}

\textbf{Calibration}
& \textbf{Evaluation}
& \textbf{AR}
& \cellcolor{tablegraylight}\textbf{ValuSpec}
& \textbf{AR}
& \cellcolor{tablegraylight}\textbf{ValuSpec} \\
\midrule

\multirow{2}{*}{HumanEval}
& GSM8K
& \makecell{23.97\\{\footnotesize (1.00$\times$)}}
& \cellcolor{tablegraylight}\makecell{
\textbf{51.94}\\
{\footnotesize \textbf{(2.17$\times$)}}
}
& \makecell{20.43\\{\footnotesize (1.00$\times$)}}
& \cellcolor{tablegraylight}\makecell{
\textbf{44.32}\\
{\footnotesize \textbf{(2.17$\times$)}}
}
\\

& Dolly
& \makecell{25.74\\{\footnotesize (1.00$\times$)}}
& \cellcolor{tablegraylight}\makecell{
\textbf{52.71}\\
{\footnotesize \textbf{(2.05$\times$)}}
}
& \makecell{17.90\\{\footnotesize (1.00$\times$)}}
& \cellcolor{tablegraylight}\makecell{
\textbf{40.03}\\
{\footnotesize \textbf{(2.24$\times$)}}
}
\\

\addlinespace[3pt]

\multirow{2}{*}{GSM8K}
& HumanEval
& \makecell{23.56\\{\footnotesize (1.00$\times$)}}
& \cellcolor{tablegraylight}\makecell{
\textbf{53.96}\\
{\footnotesize \textbf{(2.29$\times$)}}
}
& \makecell{19.43\\{\footnotesize (1.00$\times$)}}
& \cellcolor{tablegraylight}\makecell{
\textbf{45.72}\\
{\footnotesize \textbf{(2.35$\times$)}}
}
\\

& Dolly
& \makecell{25.74\\{\footnotesize (1.00$\times$)}}
& \cellcolor{tablegraylight}\makecell{
\textbf{53.65}\\
{\footnotesize \textbf{(2.08$\times$)}}
}
& \makecell{17.90\\{\footnotesize (1.00$\times$)}}
& \cellcolor{tablegraylight}\makecell{
\textbf{40.16}\\
{\footnotesize \textbf{(2.24$\times$)}}
}
\\

\addlinespace[3pt]

\multirow{2}{*}{Dolly}
& HumanEval
& \makecell{23.56\\{\footnotesize (1.00$\times$)}}
& \cellcolor{tablegraylight}\makecell{
\textbf{52.73}\\
{\footnotesize \textbf{(2.24$\times$)}}
}
& \makecell{19.43\\{\footnotesize (1.00$\times$)}}
& \cellcolor{tablegraylight}\makecell{
\textbf{44.76}\\
{\footnotesize \textbf{(2.30$\times$)}}
}
\\

& GSM8K
& \makecell{23.97\\{\footnotesize (1.00$\times$)}}
& \cellcolor{tablegraylight}\makecell{
\textbf{50.68}\\
{\footnotesize \textbf{(2.11$\times$)}}
}
& \makecell{20.43\\{\footnotesize (1.00$\times$)}}
& \cellcolor{tablegraylight}\makecell{
\textbf{42.38}\\
{\footnotesize \textbf{(2.07$\times$)}}
}
\\

\bottomrule
\end{tabular}
\end{table}

We next test whether the verifier-side valuation criterion calibrated on one workload remains effective on another without re-calibration. Specifically, $\tau_{\mathrm{tr}}$ is estimated on the 70\% calibration split of one dataset and directly applied to a different dataset, while the upstream candidate-supply procedure remains unchanged.
Table~\ref{tab:cross_dataset_results} shows that ValuSpec achieves 2.05$\times$--2.29$\times$ speedup on the 7B target and 2.07$\times$--2.35$\times$ on the 13B target across all transfer directions. Thus, the calibrated threshold captures a reusable notion of candidate verification value rather than merely fitting a particular calibration workload.

\subsection{Integration with Dynamically Constructed Trees}
\label{sec:dynamic_tree_eval}

\begin{table}[t]
\centering
\caption{\textbf{Verifier-side integration with EAGLE-3.}
Throughput (tokens/s) on Meta-Llama/Llama-3.1-8B-Instruct.
ValuSpec is applied after EAGLE-3 tree construction and before target verification, without modifying the upstream generation procedure.
}
\label{tab:eagle3_integration}

\small
\setlength{\tabcolsep}{4.2pt}
\renewcommand{\arraystretch}{1.18}

\begin{tabular}{@{}lccccc@{}}
\toprule
&
\multicolumn{3}{c}{\textbf{Throughput (tok/s)}}
&
\multicolumn{2}{c}{\textbf{Speedup}} \\
\cmidrule(lr){2-4}
\cmidrule(lr){5-6}

\textbf{Dataset}
& \textbf{AR}
& \textbf{EAGLE-3}
& \cellcolor{tablegraylight}
  \makecell{\textbf{EAGLE-3}\\[-1pt]\textbf{+ ValuSpec}}
& \makecell{\textbf{vs.}\\[-1pt]\textbf{EAGLE-3}}
& \makecell{\textbf{vs.}\\[-1pt]\textbf{AR}} \\
\midrule

HumanEval
& 35.30
& 69.10
& \cellcolor{tablegraylight}\textbf{93.28}
& 1.35$\times$
& 2.64$\times$ \\

\addlinespace[2pt]

GSM8K
& 35.30
& 70.40
& \cellcolor{tablegraylight}\textbf{98.25}
& 1.40$\times$
& 2.78$\times$ \\

\addlinespace[2pt]

Dolly
& 35.30
& 55.20
& \cellcolor{tablegraylight}\textbf{83.66}
& 1.52$\times$
& 2.37$\times$ \\

\addlinespace[2pt]

SpecBench
& 35.30
& 47.70
& \cellcolor{tablegraylight}\textbf{69.17}
& 1.45$\times$
& 1.96$\times$ \\

\bottomrule
\end{tabular}
\end{table}

EAGLE-3 dynamically constructs and optimizes its candidate tree on the generation side. We leave this procedure unchanged and invoke ValuSpec only after the tree has been materialized and before target verification. ValuSpec therefore does not participate in EAGLE-3 candidate generation, tree expansion, or generation-side pruning.
Table~\ref{tab:eagle3_integration} shows that ValuSpec still provides 1.35$\times$--1.52$\times$ additional speedup over EAGLE-3 and up to 2.78$\times$ over AR. Although EAGLE-3 already performs strong candidate shaping during tree construction, residual redundancy remains before target verification. These results show that generation-side tree optimization and verifier-side candidate valuation operate at different stages and can be composed.

\paragraph{Additional validation.}
We further evaluate this separation between candidate supply and candidate valuation using the Qwen3.5 model family. ValuSpec is applied to Standard SD, Retrieval-SD, and EAGLE-3 while leaving each upstream generation procedure unchanged. Table~\ref{tab:additional_validation} compactly summarizes the additional rebuttal experiments.
\begin{table}[t]
\centering
\caption{\textbf{Additional validation on Qwen3.5.}
Results across Standard SD, Retrieval-SD, and EAGLE-3.
Speedup ranges are across HumanEval, GSM8K, and Dolly; cross-dataset ranges summarize all transfer directions.
``Fixed'' denotes the best fixed likelihood threshold in $[0.3,0.8]$ on HumanEval.
}
\label{tab:additional_validation}

\small
\setlength{\tabcolsep}{4.0pt}
\renewcommand{\arraystretch}{1.16}

\begin{tabular}{@{}lccc@{}}
\toprule
\textbf{Metric}
& \makecell{\textbf{Standard}\\[-1pt]\textbf{SD}}
& \textbf{Retrieval-SD}
& \textbf{EAGLE-3} \\
\midrule

\rowcolor{tablegray}
\multicolumn{4}{@{}l}{\textbf{Candidate-supplier compatibility}} \\

Marginal speedup
& \textbf{1.63--1.89$\times$}
& \textbf{1.97--2.14$\times$}
& \textbf{1.38--1.61$\times$} \\

\addlinespace[2pt]

\rowcolor{tablegray}
\multicolumn{4}{@{}l}{\textbf{Cross-dataset threshold transfer}} \\

Marginal speedup
& 1.24--1.63$\times$
& 1.78--2.01$\times$
& 1.33--1.46$\times$ \\

\addlinespace[2pt]

\rowcolor{tablegray}
\multicolumn{4}{@{}l}{\textbf{Fixed-threshold comparison}} \\

Best fixed $\rightarrow$ ValuSpec
& 58.26 $\rightarrow$ \textbf{79.78}
& 87.15 $\rightarrow$ \textbf{113.16}
& 78.03 $\rightarrow$ \textbf{90.49} \\

\addlinespace[2pt]

\rowcolor{tablegray}
\multicolumn{4}{@{}l}{\textbf{Verifier-side filter behavior}} \\

Empty tree
& 0.71--1.37\%
& 0.17--0.68\%
& 0.07--0.78\% \\

Revert AR
& 1.56--2.14\%
& 1.25--1.88\%
& 1.43--3.23\% \\

False discard
& 1.35--1.91\%
& 0.75--1.58\%
& 0.13--0.44\% \\

Remaining branches
& 72.8--76.9\%
& 79.1--82.6\%
& 84.5--88.2\% \\

\addlinespace[2pt]

\rowcolor{tablegray}
\multicolumn{4}{@{}l}{\textbf{Batched inference}} \\

Batch 1 / 4 / 8
& 1.89 / 1.51 / 1.37$\times$
& 2.01 / 1.73 / 1.47$\times$
& 1.46 / 1.32 / 1.23$\times$ \\

\bottomrule
\end{tabular}
\end{table}
The same verifier-side module consistently improves all three candidate suppliers, providing 1.38$\times$--2.14$\times$ additional speedup without changing how candidates are generated. The gains persist under cross-dataset threshold transfer and batched inference. Moreover, after sweeping fixed likelihood thresholds from 0.3 to 0.8, even the best fixed cutoff remains below ValuSpec for every supplier, supporting the benefit of target-aligned calibration rather than filtering by a universal likelihood threshold.

\subsection{Acceptance, Filtered Tree Size, and Verification Efficiency}
\label{sec:mechanism_analysis}

To characterize the residual redundancy that remains after candidate generation, we analyze the candidates that reach target verification. We set the maximum candidate-tree budget to 64 for retrieval-based trees and 256 for dynamically constructed trees (EAGLE-3); the retained tree size is determined only by branch scores and the calibrated threshold $\tau_{\mathrm{tr}}$.
For EAGLE-3 without ValuSpec, acceptance rates range from 48.2\% to 71.4\% across workloads. After applying ValuSpec, the acceptance rate of the retained candidates consistently exceeds 86\%, showing that even dynamically optimized trees contain branches with limited verification value. The additional Qwen3.5 analysis in Table~\ref{tab:additional_validation} provides the same picture across candidate suppliers: ValuSpec retains approximately 73\%--88\% of supplied branches, while empty-tree rates remain below 1.4\% and false-discard rates below 2\%.
These analyses concern only the post-generation stage: the upstream candidates are fixed before ValuSpec is invoked. ValuSpec therefore reduces redundant target-side verification without changing how those candidates were originally generated. Output identity is preserved by the target model's greedy verification and autoregressive fallback rather than by the filtering decision itself.
\section{Discussion and Limitations}
\label{sec:discussion}

ValuSpec targets residual verifier-side redundancy after candidate generation has already been completed, and its benefit therefore depends on how much redundant verification remains in the candidate tree supplied by the upstream method. The amount of residual verifier-side redundancy varies across candidate suppliers, model families, and workloads; nevertheless, the consistent gains across Standard SD, Retrieval-SD, and EAGLE-3 show that generation-side candidate optimization and verifier-side candidate valuation are complementary rather than competing mechanisms.
Our current study focuses on greedy decoding with output-identical verification, and extending candidate valuation to stochastic decoding remains future work. The gain may also decrease when upstream suppliers already produce highly selective trees or when workloads contain little reusable structure. Finally, ValuSpec introduces an additional lightweight scoring step before target verification; although our batch-size experiments show that its benefits persist under batched inference, deeper integration and profiling in production serving systems remain important directions for future work.

\section{Conclusion}

We study verifier-side redundancy in tree-based speculative decoding and identify candidate valuation after candidate generation but before target verification as a distinct optimization opportunity. Unlike generation-side methods that improve how candidate sequences or trees are generated, retrieved, expanded, or pruned, ValuSpec operates on an already materialized candidate tree and determines which supplied branches are worth forwarding to the target verifier. By calibrating a filtering threshold from target-model greedy trajectories, ValuSpec provides a training-free and plug-and-play verifier-side candidate valuation mechanism that removes low-value branches before expensive target verification while preserving exact output equivalence under deterministic decoding. Experiments across model families, workloads, and candidate suppliers show that ValuSpec consistently improves verification efficiency without modifying the upstream generation procedure, including an additional 1.38$\times$--2.14$\times$ speedup over Standard SD, Retrieval-SD, and EAGLE-3 on Qwen3.5, while remaining effective under cross-dataset threshold transfer and batched inference. These results support verifier-side candidate valuation as an optimization stage complementary to generation-side candidate supply and tree optimization.

\bibliographystyle{unsrt}
\bibliography{references}

\appendix

\section{Sensitivity Analysis}
\label{sec:ablations}

\subsection{Draft Length Sensitivity}
\label{sec:sensitivity_draft}
We study the sensitivity of SD and ValuSpec to draft length $n$ by evaluating JackFram/LLaMA-68M and Llama-2-\{7b,13b\}-chat-hf on both RTX~4090 and A100. 
Table~\ref{tab:dolly_sd_ValuSpec} reports results on Dolly; corresponding results on HumanEval and GSM8K are provided in table~\ref{tab:humaneval_gsm8k_sd_ValuSpec}. 
For SD on Dolly, the acceptance rate decreases monotonically as the draft length $n$ increases, while speedup is non-monotonic and peaks at  $n{=}3$, reflecting the trade-off between verification gain and verification overhead. RTX 4090 follows the same trend but with smaller relative gains, as its higher autoregressive throughput makes SD overheads harder to amortize. 
For ValuSpec with fixed tree hyperparameters (draft tokens $=64$), the best performance consistently occurs at $n{=}4$ across A100/7b, A100/13b, and RTX~4090/7b on Dolly, and we observe the same optimal choice $n{=}4$ for HumanEval and GSM8K. 
Compared with SD, ValuSpec tolerates a slightly larger optimal $n$ because target-aligned pre-filtering suppresses low-quality branches before target verification.

\begin{table*}[b!]
\centering
\small
\setlength{\tabcolsep}{3.0pt}
\renewcommand{\arraystretch}{1.15}
\caption{Combined ablations on Dolly: SD (throughput / speedup / hit rate) across hardware/model settings, plus ValuSpec (throughput / speedup) with a fixed tree configuration (draft tokens $=64$ ). “--” indicates not applicable or not reported.}
\label{tab:dolly_sd_ValuSpec}
\begin{tabular}{llccc ccc ccc}
\toprule
Database & Method
& \multicolumn{3}{c}{A100 / 7b} & \multicolumn{3}{c}{A100 / 13b} & \multicolumn{3}{c}{RTX 4090 / 7b} \\
\cmidrule(lr){3-5}\cmidrule(lr){6-8}\cmidrule(lr){9-11}
& & Thpt & Speedup & Hit & Thpt & Speedup & Hit & Thpt & Speedup & Hit \\
\midrule

Dolly & AR
& 25.74 & -- & -- & 17.90 & -- & -- & 51.89 & -- & -- \\
Dolly & SD $n{=}1$
& 32.68 & 1.27 & 0.46 & 24.48 & 1.37 & 0.43 & 61.74 & 1.19 & 0.45 \\
Dolly & SD $n{=}2$
& 34.37 & 1.34 & 0.33 & 25.85 & 1.44 & 0.32 & 65.71 & 1.27 & 0.33 \\
Dolly & \textbf{SD $n{=}3$}
& \textbf{34.89} & \textbf{1.36} & \textbf{0.26}
& \textbf{26.99} & \textbf{1.51} & \textbf{0.22}
& \textbf{66.14} & \textbf{1.27} & \textbf{0.25} \\
Dolly & SD $n{=}4$
& 33.92 & 1.32 & 0.21 & 26.44 & 1.48 & 0.20 & 64.18 & 1.24 & 0.20 \\
Dolly & SD $n{=}5$
& 32.41 & 1.26 & 0.17 & 25.03 & 1.40 & 0.15 & 60.77 & 1.17 & 0.17 \\

\midrule

Dolly & ValuSpec $n{=}1$
& 43.78 & 1.70 & -- & 32.17 & 1.80 & -- & 65.78 & 1.27 & -- \\
Dolly & ValuSpec $n{=}2$
& 45.42 & 1.76 & -- & 33.89 & 1.89 & -- & 74.13 & 1.43 & -- \\
Dolly & ValuSpec $n{=}3$
& 48.93 & 1.90 & -- & 37.62 & 2.10 & -- & 82.59 & 1.59 & -- \\
Dolly & \textbf{ValuSpec $n{=}4$}
& \textbf{56.19} & \textbf{2.18} & --
& \textbf{41.74} & \textbf{2.33} & --
& \textbf{86.83} & \textbf{1.67} & -- \\
Dolly & ValuSpec $n{=}5$
& 49.36 & 1.92 & -- & 39.53 & 2.21 & -- & 85.66 & 1.65 & -- \\
Dolly & ValuSpec $n{=}6$
& 44.25 & 1.72 & -- & 33.68 & 1.88 & -- & 75.37 & 1.45 & -- \\

\bottomrule
\end{tabular}
\end{table*}

\begin{table*}[h]
\centering
\small
\setlength{\tabcolsep}{3.0pt}
\renewcommand{\arraystretch}{1.15}
\caption{Combined ablations on HumanEval and GSM8K: SD (throughput / speedup / hit rate) across hardware/model settings, plus ValuSpec (throughput / speedup). “--” indicates not applicable or not reported.}
\label{tab:humaneval_gsm8k_sd_ValuSpec}
\begin{tabular}{llccc ccc ccc}
\toprule
Database & Method
& \multicolumn{3}{c}{A100 / 7b} & \multicolumn{3}{c}{A100 / 13b} & \multicolumn{3}{c}{RTX 4090 / 7b} \\
\cmidrule(lr){3-5}\cmidrule(lr){6-8}\cmidrule(lr){9-11}
& & Thpt & Speedup & Hit & Thpt & Speedup & Hit & Thpt & Speedup & Hit \\
\midrule

HumanEval & AR
& 23.56 & -- & -- & 19.43 & -- & -- & 49.18 & -- & -- \\
HumanEval & SD $n{=}1$
& 27.87 & 1.18 & 0.35 & 23.26 & 1.20 & 0.34 & 57.32 & 1.17 & 0.32 \\
HumanEval & SD $n{=}2$
& 28.68 & 1.22 & 0.24 & 24.69 & 1.27 & 0.23 & 58.23 & 1.18 & 0.25 \\
HumanEval & \textbf{SD $n{=}3$}
& \textbf{29.20} & \textbf{1.24} & \textbf{0.18}
& \textbf{25.27} & \textbf{1.30} & \textbf{0.19}
& \textbf{58.29} & \textbf{1.19} & \textbf{0.18} \\
HumanEval & SD $n{=}4$
& 28.88 & 1.23 & 0.17 & 24.21 & 1.25 & 0.15 & 54.89 & 1.12 & 0.14 \\
HumanEval & SD $n{=}5$
& 24.46 & 1.04 & 0.11 & 23.08 & 1.19 & 0.11 & 53.47 & 1.09 & 0.12 \\
HumanEval & ValuSpec $n{=}4$
& 54.44 & 2.31 & -- & 46.18 & 2.38 & -- & 90.26 & 1.84 & -- \\

\midrule

GSM8K & AR
& 23.97 & -- & -- & 20.43 & -- & -- & 51.76 & -- & -- \\
GSM8K & SD $n{=}1$
& 30.51 & 1.27 & 0.47 & 25.16 & 1.23 & 0.43 & 61.97 & 1.20 & 0.44 \\
GSM8K & SD $n{=}2$
& 31.36 & 1.31 & 0.33 & 27.22 & 1.33 & 0.33 & 66.40 & 1.28 & 0.33 \\
GSM8K & \textbf{SD $n{=}3$}
& \textbf{32.07} & \textbf{1.34} & \textbf{0.26}
& \textbf{28.32} & \textbf{1.39} & \textbf{0.25}
& \textbf{67.39} & \textbf{1.30} & \textbf{0.26} \\
GSM8K & SD $n{=}4$
& 31.01 & 1.29 & 0.21 & 25.96 & 1.27 & 0.19 & 64.76 & 1.25 & 0.21 \\
GSM8K & SD $n{=}5$
& 29.88 & 1.25 & 0.17 & 24.95 & 1.22 & 0.15 & 61.11 & 1.18 & 0.17 \\
GSM8K & ValuSpec $n{=}4$
& 53.02 & 2.21 & -- & 45.76 & 2.24 & -- & 89.77 & 1.73 & -- \\

\bottomrule
\end{tabular}
\end{table*}

\subsection{Quantile Setting and Overhead}
\label{sec:sensitivity_quantile}

ValuSpec introduces a quantile hyperparameter $\delta$, which controls the conservativeness of branch filtering. Since all experiments in this paper use greedy decoding and require exact output equivalence with deterministic autoregressive decoding, we adopt a conservative setting and fix $\delta=0.05$ in all main experiments. In this way, the calibrated threshold avoids overly aggressive filtering, so that low-quality branches can be removed while preserving the candidate structure needed for output-identical verification.

Importantly, the threshold calibration of ValuSpec is performed offline. Once the transfer threshold $\tau_{\mathrm{tr}}$ is calibrated on historical workload data, it can be directly reused during inference, and does not need to be recomputed for every request. This design makes ValuSpec naturally suitable for recurring-workload settings, where similar requests repeatedly appear over time. It also supports cross-dataset transfer, where the same calibrated threshold can be applied to another workload without re-calibration.

As a result, the main additional cost of ValuSpec during deployment comes only from lightweight branch filtering, rather than repeated threshold estimation. The offline calibration is performed once and can then be reused across future requests from the same workload, and even transferred to related workloads without re-calibration. Since the valuation model is substantially smaller than the target model, the resulting online overhead remains small relative to target-side verification.

\begin{figure*}[h]
  \centering
  \includegraphics[width=0.81\linewidth]{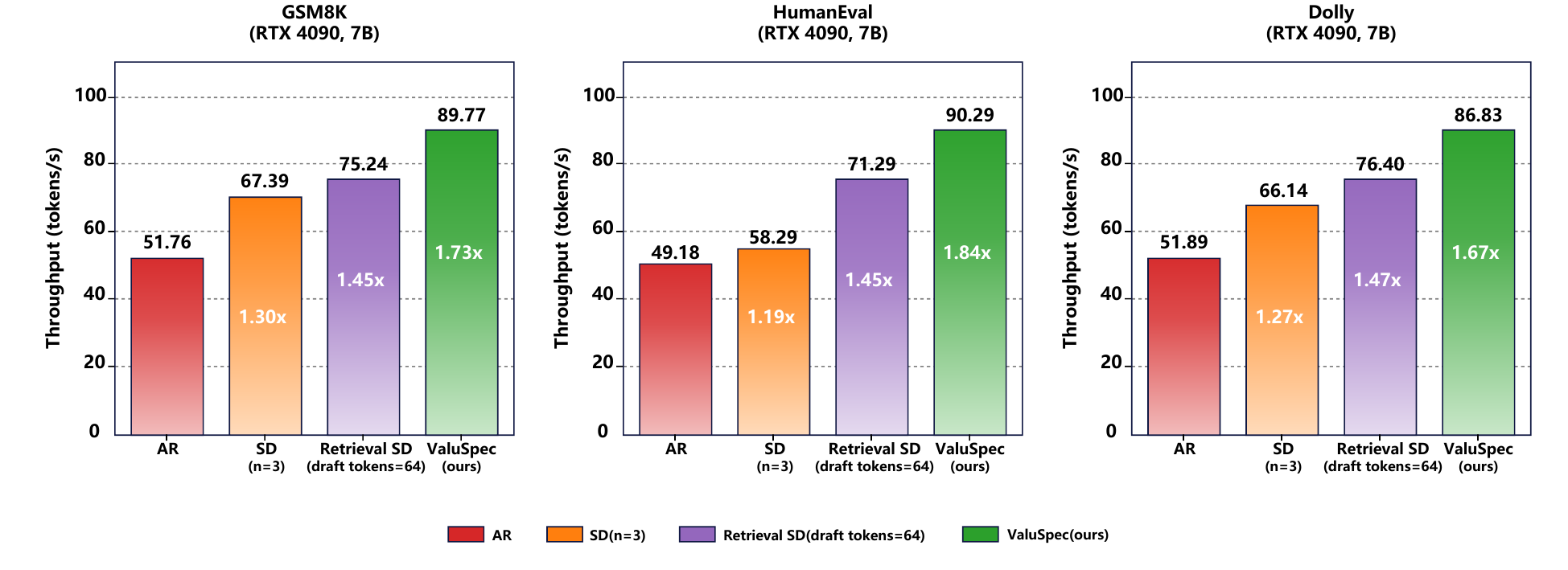}
  \caption{Cross-GPU ablations on RTX 4090. End-to-end throughput (tokens/s) and speedup over autoregressive decoding (AR) on GSM8K, HumanEval, and Dolly using Llama-2-7b-chat-hf (target) and JackFram/LLaMA-68M. We compare AR, SD ($n=3$), Retrieval SD (draft tokens $=64$), and ValuSpec (ours). Here, 7B denotes Llama-2-7b-chat-hf.}
  \label{fig:rtx4090_ablations}
  \vspace{-0.2cm}
\end{figure*}
\section{Full Algorithm of ValuSpec}
\label{app:full_algorithm}

Due to space constraints, we present the full pseudocode of ValuSpec in Algorithm~\ref{alg:ValuSpec}. The algorithm explicitly separates offline threshold calibration on $\mathcal{D}_{\mathrm{cal}}$ from online candidate-tree filtering and output-identical verification on trees constructed from $\mathcal{D}_{\mathrm{eval}}$, and details the complete decoding procedure.

\begin{algorithm}[t]
\caption{ValuSpec (offline threshold calibration on $\mathcal{D}_{\mathrm{cal}}$ + online filtering of candidate trees from $\mathcal{D}_{\mathrm{eval}}$)}
\label{alg:ValuSpec}
\small
\begin{algorithmic}[1]
\REQUIRE database $\mathcal{D}$ partitioned into disjoint splits $\mathcal{D}_{\mathrm{cal}}$ (calibration) and $\mathcal{D}_{\mathrm{eval}}$ (evaluation), with $\mathcal{D}_{\mathrm{cal}}\cap \mathcal{D}_{\mathrm{eval}}=\emptyset$
\REQUIRE a subset of $\mathcal{D}_{\mathrm{eval}}$ used to construct candidate trees, and the remaining disjoint subset used for online decoding
\REQUIRE lightweight plug-and-play candidate valuation model  $q_\phi$, target model $p_\theta$, quantile $\delta$
\ENSURE output-identical sequence $Y$

\STATE \textbf{Offline: threshold calibration on $\mathcal{D}_{\mathrm{cal}}$.}
\FOR{$x \in \mathcal{D}_{\mathrm{cal}}$}
  \STATE obtain target greedy trajectory $b^{*}(x)$ under $p_\theta$
  \STATE compute score $s_\phi(b^{*}(x)\mid x)$
\ENDFOR
\STATE $\tau_{\mathrm{tr}} \gets \mathrm{Quantile}_\delta\!\Big(\{\, s_\phi(b^{*}(x)\mid x)\;:\;x\in\mathcal{D}_{\mathrm{cal}} \,\}\Big)$

\STATE \textbf{Online: decoding using candidate trees from $\mathcal{D}_{\mathrm{eval}}$.}
\STATE initialize context $x$ and output $Y \gets [\,]$
\WHILE{not EOS}
  \STATE obtain a context-conditioned candidate tree $T_x$ constructed from $\mathcal{D}_{\mathrm{eval}}$
  \STATE filter low-quality branches:
  \[
  T_x^{\mathrm{filtered}} \gets \{\, b \in T_x : s_\phi(b\mid x)\ge\tau_{\mathrm{tr}} \,\}
  \]

  \IF{$T_x^{\mathrm{filtered}} = \emptyset$}
    \STATE $\hat y \gets \arg\max_y p_\theta(y\mid x)$
    \STATE $Y \gets Y \oplus \hat y$, \quad $x \gets x \oplus \hat y$
    \IF{$\hat y=\texttt{EOS}$}
      \STATE \textbf{break}
    \ENDIF
    \STATE \textbf{continue}
  \ENDIF

  \STATE one target forward pass with tree masking/attention on $T_x^{\mathrm{filtered}}$
  \STATE $\Delta \gets \textsc{TreeMCSD-Accept}(p_\theta, x, T_x^{\mathrm{filtered}})$

  \IF{$\Delta = [\,]$}
    \STATE $\hat y \gets \arg\max_y p_\theta(y\mid x)$
    \STATE $Y \gets Y \oplus \hat y$, \quad $x \gets x \oplus \hat y$
    \IF{$\hat y=\texttt{EOS}$}
      \STATE \textbf{break}
    \ENDIF
  \ELSE
    \STATE $Y \gets Y \oplus \Delta$, \quad $x \gets x \oplus \Delta$
    \IF{$\texttt{EOS}\in\Delta$}
      \STATE \textbf{break}
    \ENDIF
  \ENDIF
\ENDWHILE
\STATE \textbf{return} $Y$
\end{algorithmic}
\end{algorithm}

\section{Cross-hardware evaluation on RTX 4090.}
\label{SM B: 4090}
While our primary experiments use an A100 GPU, we additionally evaluate on an RTX 4090 to assess generalization across hardware platforms. As a widely accessible consumer GPU with different compute--memory trade-offs, the RTX 4090 provides a realistic deployment scenario and verifies that ValuSpec's gains are not specific to datacenter GPU characteristics.
Figure~\ref{fig:rtx4090_ablations} reports the study on RTX 4090 across GSM8K, HumanEval, and Dolly, comparing AR, SD ($n=3$), Retrieval SD (draft tokens $=64$), and ValuSpec. Across all databases, ValuSpec consistently achieves the highest throughput, indicating that target-aligned pre-filtering effectively reduces verification waste under draft--target mismatch. ValuSpec consistently achieves 1.67$\times$--1.84$\times$ speedup over AR, substantially outperforming both SD and Retrieval SD. Notably, the performance gap between Retrieval SD and ValuSpec (e.g., 1.45$\times$ vs. 1.84$\times$ on HumanEval) reveals that simply retrieving more candidate branches from the database is insufficient. ValuSpec's filtering mechanism addresses this by pruning unpromising candidates before target verification. These consistent improvements on consumer hardware confirm that ValuSpec's design is robust to different memory bandwidth and compute capabilities.

\newpage
\section*{NeurIPS Paper Checklist}

\begin{enumerate}

\item {\bf Claims}
    \item[] Question: Do the main claims made in the abstract and introduction accurately reflect the paper's contributions and scope?
    \item[] Answer: \answerYes{}
    \item[] Justification: The paper positions ValuSpec as a plug-and-play verifier-side candidate valuation module that operates after candidate generation and before target verification. This claim is supported by experiments with retrieval-based trees, Standard SD, and EAGLE-3, as well as additional validation on Qwen3.5, cross-dataset threshold transfer, fixed-threshold comparisons, filter-behavior analysis, and batched inference.

\item {\bf Limitations}
    \item[] Question: Does the paper discuss the limitations of the work performed by the authors?
    \item[] Answer: \answerYes{}
    \item[] Justification: Section~\ref{sec:discussion} discusses that ValuSpec's benefit depends on the amount of residual verifier-side redundancy left by the upstream candidate supplier. It also discusses the focus on greedy output-identical decoding, potentially smaller gains for already highly selective candidate trees or workloads with little reusable structure, the additional verifier-side scoring step, and the need for deeper production-system integration and profiling.

\item {\bf Theory assumptions and proofs}
    \item[] Question: For each theoretical result, does the paper provide the full set of assumptions and a complete proof?
    \item[] Answer: \answerNA{}
    \item[] Justification: The paper does not claim formal theoretical results. It proposes an inference-time algorithmic mechanism and evaluates it empirically.

\item {\bf Experimental result reproducibility}
    \item[] Question: Does the paper fully disclose all information needed to reproduce the main experimental results?
    \item[] Answer: \answerYes{}
    \item[] Justification: The paper specifies the datasets, model configurations, decoding setting, calibration/evaluation protocol, threshold calibration procedure, candidate-tree budgets, candidate suppliers, metrics, hardware, and the full ValuSpec algorithm. These details provide a reproducible description of the proposed method and the main experimental settings.

\item {\bf Open access to data and code}
    \item[] Question: Does the paper provide open access to the data and code?
    \item[] Answer: \answerNo{}
    \item[] Justification: All datasets, pretrained models, and benchmark resources used in the paper are publicly available. However, the implementation code for ValuSpec has not yet been publicly released.

\item {\bf Experimental setting/details}
    \item[] Question: Does the paper specify all training and test details necessary to understand the results?
    \item[] Answer: \answerYes{}
    \item[] Justification: Section~\ref{sec:exp_setup} specifies the benchmarks, 70\%/30\% calibration/evaluation protocol, cross-dataset transfer setting, model choices, candidate-tree construction, compared methods, metrics, hardware, and implementation details. The results section additionally reports validation across Standard SD, Retrieval-SD, and EAGLE-3 on Qwen3.5, fixed-threshold comparisons, filter-behavior metrics, and batch sizes 1, 4, and 8.

\item {\bf Experiment statistical significance}
    \item[] Question: Does the paper report error bars or other statistical significance information?
    \item[] Answer: \answerNo{}
    \item[] Justification: The paper reports end-to-end throughput and speedup measurements under fixed experimental configurations, but does not report error bars, confidence intervals, or formal statistical significance tests.

\item {\bf Experiments compute resources}
    \item[] Question: Does the paper provide sufficient information on compute resources?
    \item[] Answer: \answerNo{}
    \item[] Justification: The paper reports that the main experiments are conducted on a single NVIDIA A100 GPU and additionally evaluates cross-hardware generalization on an RTX 4090. However, it does not provide a complete estimate of per-experiment runtime or total GPU compute used throughout the project.

\item {\bf Code of ethics}
    \item[] Question: Does the research conform to the NeurIPS Code of Ethics?
    \item[] Answer: \answerYes{}
    \item[] Justification: The work studies inference efficiency for language models using public models and benchmarks. It does not involve human subjects, private data, sensitive personal information, or harmful data collection.

\item {\bf Broader impacts}
    \item[] Question: Does the paper discuss both potential positive and negative societal impacts of the work performed?
    \item[] Answer: \answerNo{}
    \item[] Justification: The paper focuses on a general-purpose inference-efficiency technique and does not include a separate societal-impact analysis. ValuSpec changes the computational efficiency of existing language models rather than their underlying capabilities or generated content, so its broader societal effects are indirect and largely inherited from the models and applications in which it is deployed.

\item {\bf Safeguards}
    \item[] Question: Does the paper describe safeguards for responsible release of high-risk data or models?
    \item[] Answer: \answerNA{}
    \item[] Justification: The paper does not introduce or release a new pretrained language model, high-risk dataset, or other high-risk asset. It proposes an inference-time candidate valuation mechanism evaluated with existing public models and benchmarks.

\item {\bf Licenses for existing assets}
    \item[] Question: Are existing assets properly credited and are licenses/terms mentioned?
    \item[] Answer: \answerNo{}
    \item[] Justification: The paper cites or links the public datasets, pretrained models, and benchmark resources used in the experiments, including HumanEval, GSM8K, Dolly, SpecBench, Llama-family models, Qwen3.5, and EAGLE-3. However, the manuscript does not explicitly enumerate the license name and terms for every individual existing asset.

\item {\bf New assets}
    \item[] Question: Are new assets introduced in the paper well documented?
    \item[] Answer: \answerNA{}
    \item[] Justification: The paper does not introduce or publicly release a new dataset, pretrained model, or software asset. The main contribution is an inference-time algorithmic mechanism, which is described in the paper and supplementary algorithm.

\item {\bf Crowdsourcing and research with human subjects}
    \item[] Question: Does the paper include details for crowdsourcing or human-subject experiments?
    \item[] Answer: \answerNA{}
    \item[] Justification: The paper does not involve crowdsourcing or human-subject experiments.

\item {\bf Institutional review board approvals}
    \item[] Question: Does the paper describe IRB approvals or equivalent for human-subject research?
    \item[] Answer: \answerNA{}
    \item[] Justification: The paper does not involve human-subject research, so institutional review board approval is not applicable.

\item {\bf Declaration of LLM usage}
    \item[] Question: Does the paper describe the usage of LLMs if they are an important component of the core method?
    \item[] Answer: \answerYes{}
    \item[] Justification: Large language models are central to the experimental setting. The paper specifies the target, draft, candidate-supply, and candidate-valuation models and explains their respective roles in candidate generation, verifier-side valuation, and target verification. The evaluation covers Llama-family and Qwen3.5 models as well as EAGLE-3 dynamic candidate construction.

\end{enumerate}

\end{document}